\documentclass[11pt]{article} 

\usepackage{amsmath, amssymb, amsthm}
\usepackage{graphicx}
\usepackage{bbm}
\usepackage{mdframed} 

\usepackage[margin=1in]{geometry}
\usepackage{hyperref}

\usepackage{pgf, tikz}
\usetikzlibrary{arrows, automata, positioning}
\usepackage{cleveref}
\usepackage[T1]{fontenc}
\usepackage{amsmath}
\usepackage{bm}
\usepackage{bbm}
\usepackage{mathabx}
\usepackage{authblk} 

\usepackage{amsthm,graphicx,amsmath,amsfonts,amssymb,fullpage,url,natbib}

\usepackage{accents}

\usepackage[ruled, linesnumbered]{algorithm2e}

\newcommand\blfootnote[1]{%
  \begingroup
  \renewcommand\thefootnote{}\footnote{#1}%
  \addtocounter{footnote}{-1}%
  \endgroup
}

\def\eqref#1{equation~\ref{#1}}

\def\1{\mathbf{1}}

\def\eps{{\epsilon}}

\def\ra{{\textnormal{a}}}

\newcommand{\E}{\mathbb{E}}

\newcommand{\R}{\mathbb{R}}

\newtheorem{remark}{Remark}

\newcommand{\diag}{\text{diag}}
\newcommand{\I}{\mathbf{I}}

\newcommand{\tr}{\text{Tr}}

\newcommand{\N}{\mathcal{N}}

\newmdtheoremenv[
  linecolor=black,
  linewidth=0.8pt,
  topline=true,
  bottomline=true,
  leftline=true,
  rightline=true,
  backgroundcolor=gray!8,
  skipabove=10pt,
  skipbelow=10pt,
  innertopmargin=6pt,
  innerbottommargin=6pt,
  innerrightmargin=8pt,
  innerleftmargin=8pt,
]{boxedlemma}{Lemma}

\newmdtheoremenv[
  linecolor=black,
  linewidth=0.8pt,
  topline=true,
  bottomline=true,
  leftline=true,
  rightline=true,
  backgroundcolor=gray!8,
  skipabove=10pt,
  skipbelow=10pt,
  innertopmargin=6pt,
  innerbottommargin=6pt,
  innerrightmargin=8pt,
  innerleftmargin=8pt,
]{boxedtheorem}{Theorem}

\usepackage{natbib}
\title{A Simplified Analysis of SGD for Linear Regression with Weight Averaging}

\def\bw{\bar{\mathbf{w}}}
\def\la{\langle}
\def\ra{\rangle}
\def\v{{\mathbf{v}}}
\def\x{{\mathbf{x}}}
\def\w{{\mathbf{w}}}
\def\m{{\mathbf{m}}}
\def\H{{\mathbf{H}}}
\def\A{{\mathbf{A}}}
\def\M{{\mathbf{M}}}
\def\D{{\mathbf{D}}}
\def\Q{{\mathbf{Q}}}
\def\B{{\mathbf{B}}}
\def\bias{\text{bias}}
\def\variance{\text{variance}}

\def\LLambda{{\mathbf{\Lambda}}}
\def\llambda{{\mathbf{\lambda}}}
\def\SSigma{{\bm \Sigma}}
\def\N{\mathcal{N}}

\allowdisplaybreaks

\author[*,1,2]{Alexandru Meterez}
\author[*,1,2]{Depen Morwani}
\author[1]{Costin-Andrei Oncescu}
\author[3]{Jingfeng Wu}
\author[1,2]{\\ Cengiz Pehlevan}
\author[1,2]{Sham Kakade}

\affil[1]{Harvard University}
\affil[2]{Kempner Institute at Harvard University}
\affil[3]{University of California, Berkeley}

\begin{document}
\newcommand{\alex}[1]{{\textcolor{blue}{Alex: #1}}}
\newcommand{\depen}[1]{{\textcolor{yellow}{Depen: #1}}}

\date{}
\maketitle

\begin{abstract}
    Theoretically understanding stochastic gradient descent (SGD) in overparameterized models has led to the development of several optimization algorithms that are widely used in practice today. Recent work by~\citet{zou2021benign} provides sharp rates for SGD optimization in linear regression using constant learning rate, both with and without tail iterate averaging, based on a bias-variance decomposition of the risk. In our work, we provide a simplified analysis recovering the same bias and variance bounds provided in~\citep{zou2021benign} based on simple linear algebra tools, bypassing the requirement to manipulate operators on positive semi-definite (PSD) matrices. We believe our work makes the analysis of SGD on linear regression very accessible and will be helpful in further analyzing mini-batching and learning rate scheduling, leading to improvements in the training of realistic models.
\end{abstract}

\blfootnote{\hspace{-6mm}${}^\star$: Alphabetical ordering, equal contribution. \\ Correspondence to:  \texttt{ameterez@g.harvard.edu, dmorwani@g.harvard.edu}}

\section{Problem setup and notation}
We use bolded small letters for vectors and bolded capital letters for matrices, and we use $\1$ for a vector of ones. Moreover, for any vector $\v$, we denote as $\v^k$ the operation of raising each entry of $\v$ to the power $k$, and we use $\preceq$ to denote PSD ordering. Moreover for inequalities of the form $\v_1 \leq \v_2$, we refer to an elementwise inequality between the two vectors. For a vector $\v$, we also denote the norm induced by matrix $\A$ as $\|\v\|_\A^2 = \v^\top \A \v$. We consider the eigenvalues ordered as $\lambda_{\max} = \lambda_1$.

We are going to consider the linear regression setting with Gaussian inputs and homoscedastic noise (additive/well-specified) setting. Let $\x \sim \N(0, \H)$ and $\eps \sim \N(0, \sigma^2)$ be the independent noise,  where $\x \in \R^d$ and $\eps \in \R$. Denote $y = \la \w^\star, \x \ra + \eps$ the ground truth label of $\x$. The iterate at time $t$ for SGD will be denoted by $\w_t$.  
In the case of tail averaging, we use the bar notation for the weights i.e. $\bw_t = \frac{1}{t-s} \sum_{i=s}^t \w_i$ where each $\w \in \R^d$, with $s$ denoting the iterations that we are not averaging over. We will consider $\w_0$ to be fixed. We will denote the covariance of the iterates at time $t$ as $\SSigma_t = \E [(\w_t - \w^\star)(\w_t - \w^\star)^\top]$, where the expectation is taken over the noise induced by the SGD algorithm. We will denote the eigendecomposition of $\H := \Q \LLambda \Q$. Note that $\H$ is PSD, as it is the covariance matrix of the input samples $\x$.

The mean squared error of a given weight vector $\w$ is denoted by

\[
    L(\w) = \frac{1}{2} \E [(\la \w, \x \ra - y)^2]
\]





The following document is structured as:
\begin{itemize}
    \item Section~\ref{sec:prel_thy} contains the gist for the simplified analysis to go through for computing the bias and variance bounds.
    \item Section~\ref{sec:tail_avg} contains the bias variance bounds for the case with tail averaging for the last $N$ iterates
\end{itemize}

\section{Main Technique}
\label{sec:prel_thy}
First, we will simplify the risk expression for the iterate at time $t$.
\begin{align*}
L(\w_t) 
&= \frac{1}{2}\mathbb{E}\left[((\w_t - \w^\star)^\top \x + \varepsilon)^2\right] \\
&= \frac{1}{2}\mathbb{E}\left[ (\w_t - \w^\star)^\top \x_t \x_t^\top (\w_t - \w^\star) + 2\varepsilon (\w_t - \w^\star)^\top \x_t + \varepsilon^2 \right] \\
&= \frac{1}{2} \tr(\H \SSigma_t) + \frac{1}{2} \sigma^2 \\
&= \frac{1}{2} \tr(\Q \LLambda \Q^\top \SSigma_t) + \frac{1}{2} \sigma^2 \\
&= \frac{1}{2} \tr(\LLambda \M_t) + \frac{1}{2} \sigma^2
\end{align*}

where $\M_t := \Q^\top \SSigma_t \Q$, denotes the covariance matrix of the iterates, rotated in the Hessian eigenbasis.


If we denote as $\llambda = \diag(\LLambda)$ and $\m_t = \diag(\M_t)$, then the risk only depends on the \textit{diagonal elements of $\M_t$} i.e.:

\begin{align*}
    L(\w_t) = \frac{1}{2} \tr(\LLambda \M_t) + \frac{1}{2} \sigma^2 = \frac{1}{2} \la \llambda, \m_t \ra + \frac{1}{2} \sigma^2
\end{align*}
The equation above shows that, instead of tracking the entire covariance matrix ($\SSigma_t$) as done in the previous works~\citep{zou2021benign,jain2018parallelizing}, we can track $\m_t$ for estimating the risk at time $t$. Note that a similar technique can be used in the tail averaging case - and we will expand on this in Section~\ref{sec:tail_avg}, since the risk of the averaged iterates also depends only on the vectors $\m_t$, for the times that are included in the tail averaging. Below, we will further show that for SGD, the evolution of $\m_t$ is only dependent on $\m_{t-1}$, and not on the other elements of $\M_t$. This diagonalization idea has also been used in previous literature~\citep{bordelon2021learning, wu2023finite, wu2023many}.

Specifically, \citet{wu2023finite} considered (misspecified) ReLU regression problems, in which the PSD operator methods failed to be applied directly due to certain non-symmetric matrices (as a result of the non-linearity of ReLU coupled with misspecified noise). 
They addressed this issue by applying the PSD operator method only to the diagonals of the relevant matrices \citep[Section 7]{wu2023finite}.
This idea is further extended by \citet{wu2023many} in a matrix linear regression problem (motivated by in-context learning of linear regression).
In their problem, the trainable parameters are a matrix (second-order tensor), so their covariance is a fourth-order tensor that is updated by a linear operator given by an eighth-order tensor. Such high-order tensors are hard to analyze directly; instead, \citet{wu2023many} considered the evolution of the ``diagonals'' of the fourth-order tensors (corresponding to second-order tensors), reducing the problem to bounding a sequence of matrices updated by PSD operators \citep[Section 6]{wu2023finite}.

\subsection{SGD with batch size 1}

We begin with the SGD update rule with learning rate $\eta$:
\begin{align*}
    \w_{t+1} &= \w_t - \eta \x_t\x_t^\top(\w_t - \w^\star) - \eta \eps_t \x_t \\
    \implies \w_{t+1} - \w^\star &= (\I - \eta \x_t\x_t^\top)(\w_t - \w^\star) - \eta \eps_t \x_t \\
\end{align*}

Since the risk depends on the covariance matrix $\SSigma_t$, we will write down the recursion for $\SSigma_t$.

 \begin{align*} 
(\w_{t+1} - \w^\star)(\w_{t+1} - \w^\star)^\top &= (\I - \eta \x_t \x_t^\top)(\w_t - \w^\star)(\w_t - \w^\star)^\top(\I - \eta \x_t \x_t^\top) \\
&\quad - \eta \varepsilon (\I - \eta \x_t \x_t^\top)(\w_t - \w^\star)\x_t^\top \\
&\quad - \eta \varepsilon \x_t(\w_t - \w^\star)^\top(\I - \eta \x_t \x_t^\top) \\
&\quad + \eta^2 \varepsilon^2 \x_t \x_t^\top
\end{align*}

Taking the expectation w.r.t the SGD noise process on both the sides, we get

\[ \SSigma_{t+1} = \SSigma_t - \eta \H \SSigma_t - \eta \SSigma_t \H + \eta^2\E_{\x_t}[(\x_t^\top \SSigma_t \x_t)\x_t \x_t^\top] + \eta^2 \sigma^2 \H \]



The complication arises from calculating the fourth moment term. We compute it separately: 

\begin{boxedlemma}
    \label{lem:4th_moment}
    For $\x \sim \N(0, \H)$, we have that:
    \begin{align*}
        \E_{\x}[(\x^\top \SSigma_t \x)\x \x^\top] = 2 \H \SSigma_t \H + \tr(\H \SSigma_t) \H
    \end{align*}
\end{boxedlemma}
We defer the proof of Lemma~\ref{lem:4th_moment} to Appendix~\ref{app:proofs}.

Therefore by applying Lemma~\ref{lem:4th_moment}, we get the recurrence:
\begin{align*}
    \SSigma_{t+1} &= \SSigma_t - \eta \SSigma_t \H - \eta \H \SSigma_t + 2\eta^2 \H \SSigma_t \H + \eta^2 \tr(\H \SSigma_t)\H + \eta^2 \sigma^2 \H
\end{align*}

Note that the above recurrence was also derived in various previous works including \citet{bordelon2021learning, zou2021benign, jain2018parallelizing}. Similar ideas for SGD dynamics have also been studied from a random matrix theory point of view by modelling the risk as a Volterra equation~\citep{atanasov2025two,paquette20244+} in the high dimensional case. The crucial thing that we will show below is that the recurrence for $\m_t := \diag(\Q^\top \SSigma_t \Q)$ is dependent only on $\m_{t-1}$.

Rotating both the sides of the equation above in the basis of $\H$, i.e, multiplying on the left by $\Q^\top$ and on the right by $\Q$ we get the following recurrence:

\begin{align*}
    \M_{t+1} = \M_t - \eta \M_t \LLambda - \eta \LLambda \M_t + 2\eta^2 \LLambda \M_t \LLambda + \eta^2 \tr(\LLambda \M_t) \LLambda + \eta^2 \sigma^2 \LLambda 
\end{align*}

Pushing a $\diag$ operator on both sides and rearranging the terms we get:

\begin{equation} \label{eq:main_recur}
    \m_{t+1} = \left(\I - 2 \eta \LLambda + 2\eta^2 \LLambda^2 + \eta^2 \llambda \llambda^\top \right)\m_t + \eta^2 \sigma^2 \llambda
\end{equation}

Thus, we can see that the recurrence for $\m_t$ only depends on $\m_{t-1}$. Note that the technique above also holds for general distributions with an assumption on the fourth moment operator (Assumption 2.2 in \citet{zou2021benign}), but in this note, we will restrict ourselves to Gaussian input distribution. In the next section, we will derive the precise bounds of \citet{zou2021benign} by using the recursion above, combined with the fact that the risk only depends on $\m_t$.

\section{Tail averaging bounds}
\label{sec:tail_avg}
Before proceeding with the bounds for the tail-averaged iterate, we will simplify the equations above and split the recursion in a 'bias' and 'variance' iterate.

\subsection{Bias-Variance decomposition}
Denoting: $\A = \I - 2 \eta \LLambda + 2 \eta^2 \LLambda^2 + \eta^2 \llambda \llambda^\top$, we can unroll the recursion in~\eqref{eq:main_recur} and get:


\begin{align*}
    \m_{t+1} = \A^{t+1} \m_0 + \eta^2 \sigma^2 \left(\sum_{i=0}^t \A^i\right) \llambda := \tilde{\m}_{t+1} + \bar{\m}_{t+1}
\end{align*}
where $\bias := \tilde{\m}_{t+1}$ and $\variance := \bar{\m}_{t+1}$. The nomenclature comes from the fact that the bias term can be recovered by starting the SGD algorithm from $\w_0$ in the noiseless setting (i.e. $\sigma^2 = 0$), whereas the variance term can be recovered by starting the process from $\w^\star$ (in which case $\tilde{\m}_0 = 0$), and we are stuck in a ball where the radius is controlled by the variance of the additive noise.

\begin{boxedlemma}[$\A$ bound]
    \label{lem:A_bound}
    For $\A = \I - 2 \eta \LLambda + 2\eta^2 \LLambda^2 + \eta^2 \llambda \llambda^\top$ and $\B = (\I - \eta \LLambda)^2 + 2\eta^2 \lambda \lambda^\top$, we have that $\A \leq \B$ elementwise, hence, $\A \v \le \B \v$ for any vector $\v \geq 0$.
\end{boxedlemma}
The proof is deferred to Lemma~\ref{lem:A_bound} in Appendix~\ref{app:proofs}. In general, we will work with a general $\alpha > 0$ and a matrix $\A$ given by

\[ \A := (\I - \eta \LLambda)^2 + \alpha \eta^2 \llambda \llambda^\top \]

In Appendix \ref{app:batch_size}, we show that, for a general batch size $b$, the $\alpha$ defined above varies as $2/b$. Moreover, we will assume that the learning rate $\eta \leq \frac{1}{\lambda_{max}(\H) + \alpha \tr(\H)}$. This matches the previous works \citep{jain2018parallelizing}, where at batch size $1$, the learning rate is close to $\frac{1}{\tr(\H)}$, but as batch size goes to $\infty$, learning rate becomes close to $\frac{1}{\lambda_{max}}$.


\subsection{Simplification of tail averaged loss} 
In this section we reuse some of the above calculation, but we instead compute the bias and variance bounds for the case of tail averaging for the last $N$ iterates. Namely, the risk at time $t$ becomes:

\begin{align*}
    L(\bar{\w}_{s:s+N}) = \frac{1}{2} \E\left[ \left(\frac{1}{N} \sum_{i=s}^{s+N-1} \la \w_i - \w^\star, \x_t \ra - \eps \right)^2 \right]    
\end{align*}

We begin with the following preliminary analysis, before specializing towards bounding the 2 separate terms:
\begin{align*}
L(\bw_{s:s+N}) &= \frac{1}{2} \E \left[\left( \frac{1}{N} \sum_{i=s}^{s+N-1} \langle \w_i - \w^\star, \x_t \rangle - \eps \right)^2 \right] \\
&= \frac{1}{2N^2} \underbrace{\E \left[\sum_{i=s}^{s+N-1} \x_t^\top (\w_i - \w^\star)(\w_i - \w^\star)^\top \x_t \right]}_{T_1} 
\\&\quad+ \frac{1}{N^2} \underbrace{\E \left[\sum_{i=s}^{s+N-1} \sum_{j=i+1}^{s+N-1} \x_t^\top (\w_i - \w^\star)(\w_j - \w^\star)^\top \x_t \right]}_{T_2} 
+ \frac{\sigma^2}{2} \\
&\leq \frac{1}{N^2} \E \left[\sum_{i=s}^{s+N-1} \x_t^\top (\w_i - \w^\star)(\w_i - \w^\star)^\top \x_t \right]
\\&\quad+ \frac{1}{N^2} \E \left[\sum_{i=s}^{s+N-1} \sum_{j=i+1}^{s+N-1} \x_t^\top (\w_i - \w^\star)(\w_j - \w^\star)^\top \x_t \right]
+ \frac{\sigma^2}{2} \\
&= \frac{1}{N^2} \E \left[\sum_{i=s}^{s+N-1} \sum_{j=i}^{s+N-1} \x_t^\top (\w_i - \w^\star)(\w_j - \w^\star)^\top \x_t \right] + \frac{\sigma^2}{2} \\
&= \frac{1}{\eta N^2} \left[ \sum_{i=s}^{s+N-1} \tr\left(\SSigma_i \H (\I - (\I - \eta \H)^{s+N-i}) \H^{-1} \right) \right] + \frac{\sigma^2}{2} \\
&= \frac{1}{\eta N^2} \left[ \sum_{i=s}^{s+N-1} \tr\left(\SSigma_i (\I - (\I - \eta \H)^{s+N-i}) \right) \right] + \frac{\sigma^2}{2}
\end{align*}

So if we do again the decomposition $\H = \Q \LLambda \Q^\top$, we get:
\begin{align*}
    (\I - (\I - \eta \H)^{s+N-i}) 
    &= \Q \I \Q^\top - (\Q \I \Q^\top - \Q \eta \LLambda \Q^\top)^{s+N-i} \\
    &= \Q \I \Q^\top - \Q (\I - \eta \LLambda)^{s+N-i} \Q^\top \\
    &= \Q \left(\I - (\I - \eta \LLambda)^{s+N-i}\right) \Q^\top
\end{align*}

Plugging this identity in the previous inequality and using the cyclic property of the trace, followed by the fact that we only need to consider the diagonal we obtain:

\begin{align*}
L(\bw_{s:s+N}) 
&\leq  \frac{1}{\eta N^2} \left[\sum_{i=s}^{s+N-1} \tr(\M_i (\I - (\I - \eta \LLambda)^{s+N-i}))\right] + \frac{\sigma^2}{2} \\ 
&=  \frac{1}{\eta N^2} \sum_{i=s}^{s+N-1} \langle \m_i, \1- (\1 - \eta \lambda)^{s+N-i} \rangle + \frac{\sigma^2}{2} \\
&\leq \frac{1}{\eta N^2}  \langle \sum_{i=s}^{s+N-1} \m_i, \1- (\1 - \eta \lambda)^{N} \rangle + \frac{\sigma^2}{2}
\end{align*}

Similar to before, now we need to separately bound the bias and variance terms, which make up the $\m_t$ term. We use the same notation as in the previous section.

\subsection{Bounding the bias term}
We begin by stating the main result of this section, which is the bound for the bias term. The proof is provided in the remainder of the section:
\begin{boxedtheorem}
\label{thm:bias_term_bound}
    For threshold $k^\star = \max\{j:\lambda_j \geq \frac{1}{\eta N} \}$, the bias contribution to the risk is bounded by:

    \begin{align*}
        \bias &\leq \frac{1}{\eta^2 N^2} \|(\I - \eta \LLambda)^s (\w_0 - \w^\star)\|_{\LLambda_{0:k^\star}^{-1}}^2 + 4 \|(\I - \eta \LLambda)^s (\w_0 - \w^\star)\|_{\LLambda_{k^\star:\infty}}^2 + \\&+ \frac{\alpha (\|\w_0 - \w^\star\|_{\I_{0:k^\star}}^2 + 2(s+N)\eta \|\w_0 - \w^\star\|_{\LLambda_{k^\star:\infty}}^2)}{\eta N (1 - \eta \alpha \tr(\H))} \cdot \frac{{k^\star} + 4\eta^2N^2 \sum_{j>k^\star} \lambda_j^2}{N}
    \end{align*}

\end{boxedtheorem}
We know $\tilde{\m}_i = \A \tilde{\m}_{i-1}$. Applying Lemma~\ref{lem:A_bound}, we have:
\begin{align*}
    \tilde{\m}_t \leq (\I - \eta \LLambda)^2 \tilde{\m}_{t-1} + \alpha \eta^2 \lambda \lambda^\top \tilde{\m}_{t-1}
\end{align*}

If we denote by $\D = (\I - \eta \LLambda)^2$ and by $c_t = \alpha \eta^2 \lambda^\top \tilde{\m}_{t-1}$ we have the recurrence:

\begin{align*}
    \tilde{\m}_t &= \D^{t} \tilde{\m}_0 + \sum_{i=0}^{t-1} c_{t-i} \D^i \lambda
\end{align*}

First, we provide a bound on $c_t$ in the recurrence above.

\begin{boxedlemma}[$c_t$ bound]
    \label{lem:ct_bound}
    Let $s_t = \la \tilde{\m}_t, \1 \ra$ and $r_t = \la \tilde{\m}_t, \lambda \ra$. Then, for $c_t = \alpha \eta^2 \lambda^\top \tilde{\m}_{t-1}$, we have that:
    \begin{align*}
        \sum_{t=0}^{k-1} c_t \leq \frac{\alpha \eta(s_0 - s_t)}{1 - \eta \alpha \tr(\H)}
    \end{align*}
\end{boxedlemma}

Finally, using Lemma~\ref{lem:ct_bound}, we can derive an iterate bound for the bias term:
\begin{boxedtheorem}[Bias iterate bound]
    \label{thm:bias_iterate_bound}
    For all quantities defined as before, we can upper bound the bias iterate as:
    \begin{align*}
        \tilde{\m}_t \leq \D^t \tilde{\m}_0 + \frac{\alpha \eta(s_0 - s_t)}{1 - \eta \alpha \tr(\H)} \lambda 
    \end{align*}
\end{boxedtheorem}





\paragraph{Tail averaging}
Finally, applying Lemma~\ref{thm:bias_iterate_bound}, we can write down the bound for the tail averaging case:

\begin{align*}
    \sum_{i=s}^{s+N-1} \tilde{\m}_i \leq \D \sum_{i=s}^{s+N-2} \tilde{\m}_i + \alpha \eta \lambda \frac{s_s - s_{s+N}}{1 - \eta \alpha \tr(\H)} + \tilde{\m}_s 
\end{align*}

Denote $S_{s:s+N} = \sum_{i=s}^{s+N-1} \tilde{\m}_i$, we have:

\begin{align*}
    S_{s:s+N} &\leq \D S_{s:s+N-1} + \alpha \eta \lambda \frac{s_s - s_{s+N}}{1 - \eta \alpha \tr(\H)} + \tilde{\m}_s \\
    &\leq \sum_{k=0}^{N-1} \D^k \left(\alpha \eta  \frac{s_s - s_{s+N}}{1 - \eta \alpha \tr(\H)} \lambda + \tilde{\m}_s \right) \\
    &= \sum_{k=0}^{N-1} \D^k \tilde{\m}_s +  \sum_{k=0}^{N-1} \alpha \eta  \frac{s_s - s_{s+N}}{1 - \eta \alpha \tr(\H)} \D^k \lambda
\end{align*}

In order to bound the first term, we invoke Lemma~\ref{thm:bias_iterate_bound} and get:

\begin{align*}
    \sum_{k=0}^{N-1} \D^k \tilde{\m}_s &\leq \sum_{k=0}^{N-1} \D^k \left( \D^s \tilde{\m}_0 + \frac{\alpha \eta(s_0 - s_s)}{1 - \eta \alpha \tr(\H)} \lambda \right) \\
    &= \sum_{k=0}^{N-1} \D^{k+s} \tilde{\m}_0 + \sum_{k=0}^{N-1} \frac{\alpha \eta(s_0 - s_s)}{1 - \eta \alpha \tr(\H)} \D^k \lambda 
\end{align*}

\begin{boxedlemma}[$s_0 - s_{s+N}$ bound]
    \label{lem:si_bound}
    For $s_i$ defined as before, we have that:
    \begin{align*}
        s_0 - s_{s+N} \leq \la \tilde{\m}_0, \1 - (\1 - \eta \lambda)^{2(s+N)} \ra
    \end{align*}
\end{boxedlemma}
We defer the proof of Lemma~\ref{lem:si_bound} to Appendix~\ref{app:proofs}.

Plugging back in the sum and assembling the terms we get the final bound:

\begin{align*}
    S_{s:s+N} &\leq \sum_{k=0}^{N-1} \D^{k+s} \tilde{\m}_0 + \sum_{k=0}^{N-1} \frac{\alpha \eta(s_0 - s_{s+N})}{1 - \eta \alpha \tr(\H)} \D^k \lambda \\
    &\leq \underbrace{\sum_{k=0}^{N-1} \D^{k+s} \tilde{\m}_0}_{I_1} + \underbrace{\frac{\alpha \eta\la \tilde{\m}_0, \1 - (\1 - \eta \lambda)^{2(s+N)} \ra}{1 - \eta \alpha \tr(\H)} \sum_{k=0}^{N-1}  \D^k \lambda}_{I_2} && \text{Lemma }~\ref{lem:si_bound}
\end{align*}

Now plugging this into the contribution to the risk formula, we end up with the 2 terms:

\begin{align*}
    \frac{1}{\eta N^2} \la I_1, \1 - (\1 - \eta \lambda)^N \ra &= \frac{1}{\eta N^2} \la \sum_{k=0}^{N-1} \D^{k+s} \tilde{\m}_0, \1 - (\1 - \eta \lambda)^N \ra \\
    &= \frac{1}{\eta N^2} \sum_{k=0}^{N-1}  \la \D^s \tilde{\m}_0,  \D^k (\1 - (\1 - \eta \lambda)^N) \ra \\
    &= \frac{1}{\eta N^2} \sum_{k=0}^{N-1}  \la \omega,  \D^k (\1 - (\1 - \eta \lambda)^N) \ra\\
    &= \frac{1}{\eta N^2} \la \omega,  (\I - \D^N)(\I - \D)^{-1} (\1 - (\1 - \eta \lambda)^N) \ra \\
    &=  \frac{1}{\eta N^2} \sum_j \omega_j (1 - (1 - \eta \lambda_j)^{2N})(1 - (1 - \eta \lambda_j)^N)(1 - (1 - \eta \lambda_j)^2)^{-1} \\
    &\leq  \frac{1}{\eta^2 N^2} \sum_j \frac{\omega_j}{\lambda_j} \left[1 - (1 - \eta \lambda_j)^{2N} \right]^2 \\
    &\leq \frac{1}{\eta^2 N^2} \sum_{j \leq k^\star} \frac{\omega_j}{\lambda_j} + 4 \sum_{j > k^\star} \omega_j \lambda_j \\ 
    &= \frac{1}{\eta^2 N^2} \|(\I - \eta \LLambda)^s (\w_0 - \w^\star)\|_{\LLambda_{0:k^\star}^{-1}}^2 + 4 \|(\I - \eta \LLambda)^s (\w_0 - \w^\star)\|_{\LLambda_{k^\star:\infty}}^2
\end{align*}
where $\omega = \D^s \tilde{\m}_0$ and we have defined $k^\star = \max \{k: \lambda_k \geq \frac{1}{2 \eta N} \}$.

For the contribution of the second term to the risk, we proceed in a similar fashion.

\begin{align*}
    \frac{1}{\eta N^2} \la I_2, \1 - (\1 - \eta \lambda)^N\ra &= \frac{\alpha \la \tilde{\m}_0, \1 - (\1 - \eta \lambda)^{2(s+N)} \ra}{ N^2(1 - \eta \alpha \tr(\H))} \sum_{k=0}^{N-1} \la \1- (\1 - \eta \lambda)^{N}, \D^k \lambda \ra \\
    &= \frac{\alpha \la \tilde{\m}_0, \1 - (\1 - \eta \lambda)^{2(s+N)} \ra}{ N^2(1 - \eta \alpha \tr(\H))} \la \1- (\1 - \eta \lambda)^{N}, (\I - \D^N)(\I - \D)^{-1} \lambda \ra \\
    &\leq \frac{\alpha \la \tilde{\m}_0, \1 - (\1 - \eta \lambda)^{2(s+N)} \ra}{\eta N^2(1 - \eta \alpha \tr(\H))}  \la \1- (\1 - \eta \lambda)^{2N}, \1 - (\1 - \eta \lambda)^{2N} \ra \\
    &\leq \frac{\alpha \la \tilde{\m}_0, \1 - (\1 - \eta \lambda)^{2(s+N)} \ra}{\eta (1 - \eta \alpha \tr(\H))} \left(\frac{k^\star}{N^2} + 4\eta^2 \sum_{j>k^\star} \lambda_j^2 \right)\\
    &\leq \frac{\alpha (\|\w_0 - \w^\star\|_{\I_{0:k^\star}}^2 + 2(s+N)\eta \|\w_0 - \w^\star\|_{\LLambda_{k^\star:\infty}}^2)}{\eta (1 - \eta \alpha \tr(\H))} \left(\frac{k^\star}{N^2} + 4\eta^2 \sum_{j>k^\star} \lambda_j^2 \right)
\end{align*}

where again we have split at the same $k^\star$ as before, and bounded as:

\begin{align*}
\la \1- (\1 - \eta \lambda)^{2N}, \1 - (\1 - \eta \lambda)^{2N} \ra &\leq k^\star + 4N^2 \eta^2 \sum_{j > k^\star} \lambda_j^2\\
\la \tilde{\m}_0, \1 - (\1 - \eta \lambda)^{2(s+N)} \ra &\leq \sum_{j \leq k^\star} \tilde{\m}_0^j + \sum_{j > k^\star} \tilde{\m}_0^j 2(s+N)\eta \lambda_j \\
    &= \|\w_0 - \w^\star\|_{\I_{0:k^\star}}^2 + 2(s+N)\eta \|\w_0 - \w^\star\|_{\LLambda_{k^\star:\infty}}^2
\end{align*}

\subsection{Bounding the variance term}
As before, we begin this section by stating the main result, which we further prove in the next section.
\begin{boxedtheorem}
    For thresholds $k^\star = \max\{j:\lambda_j \geq \frac{1}{\eta N} \}$ and $k^\dagger = \max\{j:\lambda_j \geq \frac{1}{\eta (s+N)} \}$, the variance contribution to the risk is bounded by:
    \begin{align*}
        \variance \leq \frac{\sigma^2}{1 - \eta \alpha \tr(\H)} \left(\frac{k^\star}{N} + 4\eta \sum_{k^\star < j \leq k^\dagger} \lambda_j + 16 \eta^2 (s+N) \sum_{j \geq k^\dagger} \lambda_j \right)
    \end{align*}
\end{boxedtheorem}
\begin{remark} For general batch size $b$, the bound stated above further scales down as $1/b$ as explained in Appendix~\ref{app:batch_size}.
\end{remark}
First note that the recursion for the variance term is:
\begin{align*}
    \bar{\m}_t = \eta^2 \sigma^2 \sum_{i=0}^{t-1} \A^i \lambda
\end{align*}

\begin{remark}
    Note that while an explicit variance iterate bound is not necessary to conclude this proof, we do provide such a bound in Lemma~\ref{thm:variance_bound}.
\end{remark}

In order to compute the variance contribution to the risk, we need the following intermediate result:
\begin{align*}
    (\I - \A)^{-1} \lambda &\leq (\I - \B)^{-1} \lambda \\
    &\leq \frac{1}{\eta}(\LLambda - \alpha \eta \lambda \lambda^\top)^{-1} \lambda \\
    &= \frac{1}{\eta} \frac{1}{1 - \eta \alpha \tr(\H)} \1 && \text{Sherman–Morrison formula}
\end{align*}

Thus, the variance contribution to the risk becomes:
\begin{align*}
    \frac{1}{\eta N^2} \la \sum_{i=s}^{s+N-1} \bar{\m}_i, \1 - (\1 - \eta \lambda)^{N} \ra 
    &= \frac{\eta \sigma^2}{N^2} \la \sum_{i=s}^{s+N-1} \sum_{j=0}^i \A^j \lambda, \1 - (\1 - \eta \lambda)^N \ra \\
    &= \frac{\eta \sigma^2}{N^2} \la \sum_{i=s}^{s+N-1} (\I - \A^{i+1})(\I - \A)^{-1} \lambda, \1 - (\1 - \eta \lambda)^{N} \ra \\
    &\leq \frac{\sigma^2}{N^2 (1 - \eta \alpha \tr(\H))} \la \sum_{i=s}^{s+N-1} (\I - \A^{i+1}) \1, \1 - (\1 - \eta \lambda)^{N} \ra \\
    &\leq \frac{\sigma^2}{N^2(1 - \eta \alpha \tr(\H))} \la \sum_{i=s}^{s+N-1} \1 - (\1 - 2\eta \lambda)^{2(i+1)}, \1 - (\1 - \eta \lambda)^{N} \ra \\
    &\leq \frac{\sigma^2}{N(1 - \eta \alpha \tr(\H))} \la \1 - (\1 - 2 \eta \lambda)^{2(s+N)}, \1 - (\1 - 2\eta \lambda)^{2N} \ra \\
\end{align*}

Similar to~\cite{zou2021benign}, we split into 3 eigenvalue bands, at thresholds $k^\star = \max\{j:\lambda_j \geq \frac{1}{\eta N} \}$ and $k^\dagger = \max\{j:\lambda_j \geq \frac{1}{\eta (s+N)} \}$. Based on this, we can bound the dot product above as:

\begin{align*}
    (\1 - (\1 - 2 \eta \lambda)^{2(s+N)})(\1 - (\1 - 2\eta \lambda)^{2N}) &\leq 1 && \lambda_i \geq \frac{1}{\eta N} \\
    (\1 - (\1 - 2 \eta \lambda)^{2(s+N)})(\1 - (\1 - 2\eta \lambda)^{2N}) &\leq 4 \cdot \eta N \lambda_i &&  \frac{1}{\eta (s+N)} \leq \lambda_i < \frac{1}{\eta N} \\
    (\1 - (\1 - 2 \eta \lambda)^{2(s+N)})(\1 - (\1 - 2\eta \lambda)^{2N}) &\leq 16 \eta (s+N) \lambda_i \cdot \eta N \lambda_i && \lambda_i <  \frac{1}{\eta (s+N)}  \\
\end{align*}

Using these bounds we get

\[ \frac{1}{\eta N^2} \la \sum_{i=s}^{s+N-1} \bar{\m}_i, \1 - (\1 - \eta \lambda)^{N} \ra \leq \frac{\sigma^2}{1 - \eta \alpha \tr(\H)} \left(\frac{k^\star}{N} + 4\eta \sum_{k^\star < j \leq k^\dagger} \lambda_j + 16 \eta^2 (s+N) \sum_{j \geq k^\dagger} \lambda_j \right) \]

\section{Discussion}
In this note we have provided a mathematically simplified technique of deriving existing SGD rates in the optimization literature~\citep{jain2018parallelizing,zou2021benign}. While previous works make use of operators on PSD matrices tracking the whole iterate covariance, our derivation shows that it suffices to only track the diagonal of this matrix (rotated in the eigenbasis of the data), thus reducing the complexity of the analysis to simple linear algebra. However, while we recover the upper bound from~\citet{zou2021benign}, this upper bound is not sharp: namely, ~\citet{zou2021benign} prove a tighter lower bound for tailed weight averaging (provided in Appendix \ref{app:lower_bound} for completeness) and conjecture that the upper bound should also be improvable. We hope that our simplified framework will inspire future work to tackle this improvement.
Furthermore, our analysis can be extended to various learning rate scheduler schemes, thus possibly leading to improvements in the optimization of realistic models.

\section*{Acknowledgements}
AM thanks Blake Bordelon and Jacob Zavatone-Veth for insightful discussions. AM, DM acknowledges the support of a Kempner Institute Graduate Research Fellowship. CP is supported by NSF grant DMS-2134157, NSF CAREER Award IIS-2239780, DARPA grant DIAL-FP-038, a Sloan Research Fellowship, and The
William F. Milton Fund from Harvard University. AM, SK and DM acknowledge that this work has been made possible in part by a
gift from the Chan Zuckerberg Initiative Foundation to establish the Kempner Institute for the Study of Natural and Artificial Intelligence.

\newpage

\appendix
\section{Additional proofs}
\label{app:proofs}
\begin{proof}[Proof of Lemma~\ref{lem:4th_moment}]
     Since $\mathbf{P} = (\w_t - \w^\star)(\w_t - \w^\star)^\top$ is independent of $\x_t$, we can take the expectation conditioned on $\mathbf{P}$, then apply tower rule. For any centered Gaussian vector, Iserlis theorem gives us that:
    \begin{align*}
        \E [\x_i \x_k \x_l \x_j] = \H_{ik}\H_{lj} + \H_{il}\H_{kj} + \H_{ij} \H_{kl}
    \end{align*}
    Thus we get that (skipping the $t$ index for $\x_t$ to simplify notation):

    \begin{align*}
        \E [\x \x^\top \mathbf{P} \x \x^\top]_{ij} &= \sum_{k, l} \mathbf{P}_{kl} (\H_{ik}\H_{lj} + \H_{il}\H_{kj} + \H_{ij} \H_{kl}) \\
        &= 2 (\H \mathbf{P} \H)_{ij} + \tr(\H \mathbf{P}) \H_{ij}
    \end{align*}

    Reassembling and applying tower rule we get:
    \begin{align*}
        \E [\x_t \x_t^\top (\w_t - \w^\star)(\w_t - \w^\star)^\top \x_t \x_t^\top] = 2 \H \SSigma_t \H + \tr(\H \SSigma_t) \H
    \end{align*}
\end{proof}

\begin{proof}[Proof of Lemma~\ref{lem:A_bound}]
    We can upper bound $\A$ as:
    \begin{align*}
        \A &= (\I - \eta \LLambda)^2 + \eta^2(\LLambda^2 + \lambda \lambda^\top) \\
        &\leq (\I - \eta \LLambda)^2 + \eta^2(\lambda \lambda^\top + \lambda \lambda^\top) && \LLambda^2 \leq \lambda \lambda^\top \\
        &= \B
    \end{align*}
    where in the first inequality we have upper bounded $\LLambda^2$ elementwise since each $\lambda_i \geq 0$, being the eigenvalues of the data covariance matrix which is PSD.
\end{proof}

\begin{proof}[Proof of Lemma~\ref{lem:ct_bound}.]
    We start from:
    \begin{align*}
        \tilde{\m}_t \leq (\I - \eta \LLambda)^2 \tilde{\m}_{t-1} + \alpha \eta^2 \lambda \lambda^\top \tilde{\m}_{t-1} &\leq (\I - \eta \LLambda) \tilde{\m}_{t-1} + \alpha \eta^2 \lambda \lambda^\top \tilde{\m}_{t-1} \\&= \tilde{\m}_{t-1} - \eta \LLambda \tilde{\m}_{t-1} + \alpha \eta^2 \lambda \lambda^\top \tilde{\m}_{t-1}
    \end{align*}
    
    Now doing a dot product with $\1$ we get:
    \begin{align*}
        s_t \leq s_{t-1} - \eta r_{t-1} + \alpha \eta^2 \tr(\H) r_{t-1} 
        \implies \sum_{t=0}^{k-1} r_t \leq \frac{s_0 - s_k}{\eta - \eta^2 \alpha \tr(\H)} \implies \sum_{t=0}^{k-1} c_t \leq \frac{\alpha \eta(s_0 - s_t)}{1 - \eta \alpha \tr(\H)}
    \end{align*}
\end{proof}

\begin{proof}[Proof of Theorem~\ref{thm:bias_iterate_bound}.]
Based on the recursion on $\tilde{\m}_t$ we have:
    \begin{align*}
    \tilde{\m}_t &= \D^t \tilde{\m}_0 + \sum_{i=0}^{t-1} c_{t-i} \D^i \lambda \\
    &\leq \D^t \tilde{\m}_0 + \sum_{i=0}^{t-1} c_i \lambda  && \D \leq \I\\
    &\leq \D^t \tilde{\m}_0 + \frac{\alpha \eta(s_0 - s_t)}{1 - \eta \alpha \tr(\H)} \lambda && \text{Lemma }~\ref{lem:ct_bound}
    \end{align*}
\end{proof}

\begin{proof}[Proof of Lemma~\ref{lem:si_bound}.]
    First, note that $\A \geq \D$ elementwise. Thus, we can bound the following quantity:
    \begin{align*}
        \tilde{\m}_0 - \tilde{\m}_{t} &= \tilde{\m}_0 - \A^{t} \tilde{\m}_0 \leq \tilde{\m}_0 - \D^t \tilde{\m}_0 = (\I - \D^t) \tilde{\m}_0
    \end{align*}
    
    Using this, we get:
    \begin{align*}
        s_0 - s_{s+N} &\leq \la (\I - \D^t) \tilde{\m}_0, \1 \ra = \la \tilde{\m}_0, \1 - (\1 - \eta \lambda)^{2(s+N)} \ra
    \end{align*}
\end{proof}

\section{General batch size result} \label{app:batch_size}
For a general batch size $b$, we can show that the following lemma holds:

\begin{boxedlemma}
\label{lem:gen_batch}
For SGD with a general batch size $b$, $\m_t$ satisfies the following recurrence
\begin{equation} 
    \m_{t+1} = \left[\I - 2 \eta \LLambda + \eta^2 \left(1+\frac{1}{b} \right) \LLambda^2 + \frac{\eta^2}{b} \llambda \llambda^\top \right]\m_t + \frac{\eta^2}{b} \sigma^2 \llambda
\end{equation}
\end{boxedlemma}

The matrix $\A_b := \left[\I - 2 \eta \LLambda + \eta^2 \left(1+\frac{1}{b} \right) \LLambda^2 + \frac{\eta^2}{b} \llambda \llambda^\top \right]$ is upper bounded by $(\I - \eta \LLambda)^2 + \frac{2}{b} \eta^2 \llambda \llambda^\top$, thus falling into our general framework with $\alpha = \frac{2}{b}$. Moreover, we can also see that the variance term recurrence scales down as $\frac{1}{b}$, leading to reduced variance with increasing batch size.

\begin{proof}[Proof of Lemma~\ref{lem:gen_batch}.]
    In the case of minibatching, the update equation is:
\begin{align*}
    \w_{t+1} - \w^\star &= (\I - \frac{\eta}{b} \sum_{i=1}^b \x_i\x_i^\top)(\w_t - \w^\star) - \frac{\eta}{b} \eps \sum_{i=1}^b \x_i
\end{align*}

Since the risk depends on the covariance matrix $\SSigma_t$, we will write down the recursion for $\SSigma_t$. Note that the cross terms that have an $\eps$ factor will simplify in expectation, since $\eps$ is independent from the other terms. Thus, we have:

\begin{align*}
    \E [(\w_{t+1} - \w^\star)(\w_{t+1} - \w^\star)^\top] &= \E\left[  (\I - \frac{\eta}{b} \sum_{i=1}^b \x_i\x_i^\top)(\w_t - \w^\star)(\w_t - \w^\star)^\top (\I - \frac{\eta}{b} \sum_{i=1}^b \x_i\x_i^\top)^\top \right] \\
    &+\frac{\eta^2}{b^2} \sigma^2 \sum_{i, j=1}^b \E [\x_i \x_j] \\
    \implies \SSigma_{t+1} &= \SSigma_t - \eta \SSigma_t \H - \eta \H \SSigma_t + \frac{\eta^2}{b^2} \sum_{i, j=1}^b \E \left[\x_i \x_i^\top (\w_t - \w^\star)(\w_t - \w^\star)^\top \x_j \x_j^\top \right]
\end{align*}

Similar to the previous case, we need to compute a 4th moment term in the update of $\SSigma_t$. We compute it below:

 \begin{align*} 
\frac{\eta^2}{b^2} \sum_{i, j=1}^b \E \left[\x_i \x_i^\top (\w_t - \w^\star)(\w_t - \w^\star)^\top \x_j \x_j^\top \right] &=  
\frac{\eta^2}{b} \E \left[\x_i \x_i^\top (\w_t - \w^\star)(\w_t - \w^\star)^\top \x_i \x_i^\top \right] \\ &+ \frac{\eta^2 b(b-1)}{b^2} \E \left[\x_i \x_i^\top (\w_t - \w^\star)(\w_t - \w^\star)^\top \x_j \x_j^\top \right]\\
&= \frac{\eta^2}{b} (2\H \SSigma_t \H + \tr(\H \SSigma_t) \H) + \frac{\eta^2 (b-1)}{b} \H \SSigma_t \H \\
&= \frac{\eta^2}{b} \left[(b+1) \H \SSigma_t \H + \tr(\H \SSigma_t)\H \right]
\end{align*}
where the first term was computed in the proof of Lemma~\ref{lem:4th_moment}, and the second term comes from factorizing the expectation across the factors since they are independent. 

Plugging back in the previous formula for the update of $\SSigma_t$ we thus get the final formula:
\begin{align*}
    \SSigma_{t+1} &= \SSigma_t - \eta \SSigma_t \H - \eta \H \SSigma_t + \eta^2 \left(1+ \frac{1}{b}\right) \H \SSigma_t \H + \frac{\eta^2}{b} \tr(\H \SSigma_t)\H + \frac{\eta^2}{b} \sigma^2 \I
\end{align*}

Multiplying on the left by $\Q$ and on the right by $\Q^\top$ we obtain the recurrence for $\M_t$:

\begin{align*}
    \M_{t+1} = \M_t - \eta \M_t \LLambda - \eta \LLambda \M_t + \eta^2\left(1 + \frac{1}{b}\right) \LLambda \M_t \LLambda + \frac{\eta^2}{b} \tr(\LLambda \M_t) \LLambda + \frac{\eta^2}{b} \sigma^2 \I
\end{align*}

Finally, pushing a $\diag$ operator through the whole equation yields the final $\A_b$ expression.
\end{proof}

\section{Variance iterate bound}
In this section we provide a short lemma, giving us a variance iterate bound:

\begin{boxedtheorem}
    \label{thm:variance_bound}
    For all quantities defined as before, we can bound the variance iterate as:
    \begin{align*}
        \bar{\m}_t \leq \frac{\eta \sigma^2}{1 - \eta \alpha \tr(\H)} (\1 - (\1 - \eta \lambda)^{2t})
    \end{align*}
\end{boxedtheorem}

The proof is provided in the remainder of this section. We start with:
\begin{align*}
    \bar{\m}_t = \eta^2 \sigma^2 \sum_{i=0}^{t-1} \A^i \lambda
\end{align*}
Now note that since $\A = (\I - \eta \LLambda)^2 + \alpha \eta^2 \lambda \lambda^\top \geq 0$, we have that $0 \leq \bar{\m}_t \leq \bar{\m}_{t+1}$ for all $t$. Taking $t \to \infty$ we have that:

\begin{align*}
    \bar{\m}_\infty = \eta^2 \sigma^2 (\I - \A)^{-1} \lambda \leq \frac{\eta \sigma^2}{1 - \eta \alpha \tr(\H)} \1
\end{align*}

Using this, we have that:
\begin{align*}
    \bar{\m}_t &= \eta^2 \sigma^2 \sum_{i=0}^{t-1} \A^i \lambda  \\
    &= \A \eta^2 \sigma^2 \sum_{i=0}^{t-1} \A^{i-1} \lambda \\
    &= \A \eta^2 \sigma^2 \sum_{i=-1}^{t-2} \A^i \lambda \\
    &=  \A (\bar{\m}_{t-1} + \eta^2 \sigma^2 \A^{-1} \lambda) \\
    &= \A \bar{\m}_{t-1} + \eta^2 \sigma^2  \lambda \\
    &= \D \bar{\m}_{t-1} + \alpha \eta^2 \lambda \lambda^\top \bar{\m}_{t-1} + \eta^2 \sigma^2  \lambda \\
    &\leq \D \bar{\m}_{t-1} + \alpha \eta^2 \tr(\H) \frac{\eta \sigma^2}{1 - \eta \alpha \tr(\H)} \lambda + \eta^2 \sigma^2 \lambda \\
    &= \D \bar{\m}_{t-1} + \eta^2 \sigma^2 \left( \frac{\alpha \tr(\H) \eta}{1 - \eta \alpha \tr(\H)} + 1 \right) \lambda \\
    &= \D \bar{\m}_{t-1} + \frac{\eta^2 \sigma^2}{1 - \eta \alpha \tr(\H)} \lambda \\
    &= \D^{t} \bar{\m}_0 + \frac{\eta^2 \sigma^2}{1 - \eta \alpha \tr(\H)} \sum_{i=0}^{t-1} \D^i \lambda
\end{align*}

Now unrolling this recursion and using the fact that $\bar{\m}_0 = 0$, we get:

\begin{align*}
    \bar{\m}_t &\leq \D^{t} \bar{\m}_0 + \frac{\eta^2 \sigma^2}{1 - \eta \alpha \tr(\H)} \sum_{i=0}^{t-1} \D^i \lambda \\
    &=\D^{t} \bar{\m}_0 + \frac{\eta^2 \sigma^2}{1 - \eta \alpha \tr(\H)} (\I - \D^{t})(\I - \D)^{-1} \lambda \\
    &\leq \D^{t} \bar{\m}_0 + \frac{\eta \sigma^2}{1 - \eta \alpha \tr(\H)} (\1 - (\1 - \eta \lambda)^{2t}) \\
    &= \frac{\eta \sigma^2}{1 - \eta \alpha \tr(\H)} (\1 - (\1 - \eta \lambda)^{2t})
\end{align*}

\section{Lower bounds} \label{app:lower_bound}
While we do not cover lower bounds in this note, the work by~\citet{zou2021benign} provides a lower bound that is sharper than the current upper bound. For sake of completeness, we will first restate (without proof) the lower bound from~\citet{zou2021benign} in our mathematical framework:

\begin{boxedtheorem}[Theorem 5.2 from~\citet{zou2021benign}]
    For SGD with tail averaging and thresholds $k^\star = \max\{j:\lambda_j \geq \frac{1}{\eta N} \}$ and $k^\dagger = \max\{j:\lambda_j \geq \frac{1}{\eta (s+N)} \}$, the bias and variance are lower bounded by (under mild assumptions):
    \begin{align*}
        \bias &\gtrsim \frac{1}{\eta^2 N^2} \|(\I - \eta \LLambda)^s (\w_0 - \w^\star)\|_{\LLambda_{0:k^\star}^{-1}}^2 + \|(\I - \eta \LLambda)^s (\w_0 - \w^\star)\|_{\LLambda_{k^\star:\infty}}^2 \\
        &+ \|\w_0 - \w^\star \|_{\LLambda_{k^\dagger:\infty}}^2\left(\frac{k^\star}{N} + N\eta^2 \sum_{j>k^\star} \lambda_j^2 \right)\\
        \variance &\gtrsim \sigma^2 \left(\frac{k^\star}{N} + \eta \sum_{k^\star < j \leq k^\dagger} \lambda_j + (s+N) \eta^2 \sum_{j>k^\dagger} \lambda_j^2 \right)
    \end{align*}
\end{boxedtheorem}

Note that the theorem is stated in a simplified way, ignoring the constants and focusing only on the leading order terms. The extra term $\|\w_0 - \w^\star\|_{\I_{0:k^\dagger}}^2$ that appears in the upper bound is not present in the lower bound, thus showing that the upper bound is not sharp.

\bibliography{references}
\end{document}